\documentclass[11bpt]{elsarticle}

\usepackage{epsfig}
\usepackage{graphicx}
\usepackage{amsmath}
\usepackage{amssymb}
\usepackage{bm}
\usepackage{tabularx}
\usepackage{caption}
\usepackage{threeparttable}
\usepackage{booktabs}

\usepackage{lineno}

\usepackage{algorithm}
\usepackage{algpseudocode}

\let\OLDthebibliography\thebibliography
\renewcommand\thebibliography[1]{
  \OLDthebibliography{#1}
  \setlength{\parskip}{0pt}
  \setlength{\itemsep}{0pt plus 0.3ex}
}

\makeatletter
\def\BState{\State\hskip-\ALG@thistlm}
\makeatother

\algnewcommand\algorithmicinput{\textbf{INPUT:}}
\algnewcommand\INPUT{\item[\algorithmicinput]}

\renewcommand{\vec}[1]{\text{\boldmath$#1$}} 

\DeclareMathOperator*{\argmax}{\arg\!\max}

\algnewcommand\algorithmicoutput{\textbf{OUTPUT:}}
\algnewcommand\OUTPUT{\item[\algorithmicoutput]}

\newcommand{\R}{\mathbb{R}}
\newcommand{\N}{\mathcal{N}}
\newcommand{\Sp}{\mathcal{S}}

\newcommand{\xdownarrow}[1]{%
  {\left\downarrow\vbox to #1{}\right.\kern-\nulldelimiterspace}
}

\newcommand{\xuparrow}[1]{%
  {\left\uparrow\vbox to #1{}\right.\kern-\nulldelimiterspace}
}

\newcommand\norm[1]{\left\lVert#1\right\rVert}

\makeatletter
\newcommand{\multiline}[1]{%
  \begin{tabularx}{\dimexpr\linewidth-\ALG@thistlm}[t]{@{}X@{}}
    #1
  \end{tabularx}
}
\makeatother

\begin{document}
\journal{Physics in Medicine and Biology.}

\title{Deep Learning with Cinematic Rendering: \\Fine-Tuning Deep Neural Networks Using Photorealistic Medical Images}

\author{Faisal Mahmood$^1$, Richard Chen$^2$, Sandra Sudarsky$^3$, \\Daphne Yu$^3$ and Nicholas J. Durr$^1$ \\ \{faisalm,ndurr\}@jhu.edu}

\begin{frontmatter}

\begin{abstract}
Deep learning has emerged as a powerful artificial intelligence tool to interpret medical images for a growing variety of applications. However, the paucity of medical imaging data with high-quality annotations that is necessary for training such methods ultimately limits their performance. Medical data is challenging to acquire due to privacy issues, shortage of experts available for annotation, limited representation of rare conditions and cost. This problem has previously been addressed by using synthetically generated data. However, networks trained on synthetic data often fail to generalize to real data. Cinematic rendering simulates the propagation and interaction of light passing through tissue models reconstructed from CT data, enabling the generation of photorealistic images. {In this paper, we present one of the first applications of cinematic rendering in deep learning, in which we propose to fine-tune synthetic data-driven networks using cinematically rendered CT data for the task of monocular depth estimation in endoscopy. Our experiments demonstrate that: (a) Convolutional Neural Networks (CNNs) trained on synthetic data and fine-tuned on photorealistic cinematically rendered data adapt better to real medical images and demonstrate more robust performance when compared to networks with no fine-tuning, (b) these fine-tuned networks require less training data to converge to an optimal solution, and (c) fine-tuning with data from a variety of photorealistic rendering conditions of the same scene prevents the network from learning patient-specific information and aids in generalizability of the model.} Our empirical evaluation demonstrates that networks fine-tuned with cinematically rendered data predict depth with 56.87\% less error for rendered endoscopy images and 27.49\% less error for real porcine colon endoscopy images.
\\


\end{abstract}

\end{frontmatter}

%
%
%
%

\section{Introduction}

{
Convolutional Neural Networks (CNNs) have revolutionized the fields of computer vision, machine and automation, achieving remarkable performance on previously-difficult tasks such as image classification, semantic segmentation, and depth estimation. \citep{shin2016deep,greenspan2016guest,shen2017deep,zhang2017deep}. CNNs are particularly powerful in supervised learning tasks where it is difficult to build an accurate mathematical model for the task at hand. With recent improvements made in training CNNs such as utilizing dropout regularization, skip connections and the advancements made in high-performance computing due to graphical processing units \citep{lecun2015deep,goodfellow2016deep}, deep learning models have become much easier to train and vastly more accessible.

To achieve generalization, deep learning models require large amounts of data that are accurately annotated. Obtaining such a dataset for a variety of medical images is challenging because expert annotation can be expensive, time consuming \citep{gur2017towards,moradi2016cross}, and often limited by the subjective interpretation \citep{kerkhof2007dysplasia}. Moreover, other issues such as privacy and under-representation of rare conditions impede developing such datasets \citep{wong2017building,schlegl2017unsupervised}. This is supplemented with the cross-patient adaptability problem, where networks trained on data from one patient fail to adapt to another patient \citep{reiter2016endoscopic,mahmood2017unsupervised}. For medical diagnostics, physicians are interested in diagnostic information which is common across patients rather than patient-specific information.
}

\subsection{Training with Synthetic Medical Images}
{
Recently, the limited availability of medical data has been addressed by the use of synthetic data \citep{mahmood2017deep,mahmood2017unsupervised,mahmood2018deep}. Computer graphics engines such as Blender and Unreal have the ability to construct realistic virtual worlds, but are limited by the diversity of 3D assets to create accurate, tissue-equivalent models \citep{zhang2016unreal}. Other methods for synthetic data generation include Generative Adversarial Networks (GANs), which train a generative deep network to learn and sample a target distribution of realistic images \citep{goodfellow2014generative}. This approach, however, suffers from the mode collapse problem, a commonly encountered failure case in GANs where the support size of the learned distribution is low, and thus, the generated images are sampled with low variability \citep{creswell2017gan}. Overall, networks trained on synthetic data often fail to generalize to real data, as both of these approaches in synthetic data generation fail to produce realistic, diverse examples necessary for training deep networks in medical images \citep{mahmood2017unsupervised}. Cinematic rendering is a recently developed visualization technique that works by simulating the propagation and interaction of light passing through tissue models reconstructed from cross-sectional images such as CT, enabling the generation of photorealistic images that have not previously been possible \citep{eid2017cr}. In this paper, we use cinematic rendering to generate a wide range of healthy to pathologic colon tissue with ground truth depth, and fine-tune synthetic data-driven networks with these images to address the problem of the cross-patient adaptability in training deep networks. To our knowledge, this is the first application of cinematic rendering in deep learning for medical image analysis. We test this method for monocular depth estimation in endoscopy, a task with many clinical applications \citep{durr2014colon} but is challenging to acquire accurately.
}

\begin{figure}
\centering
\includegraphics[width=\textwidth]{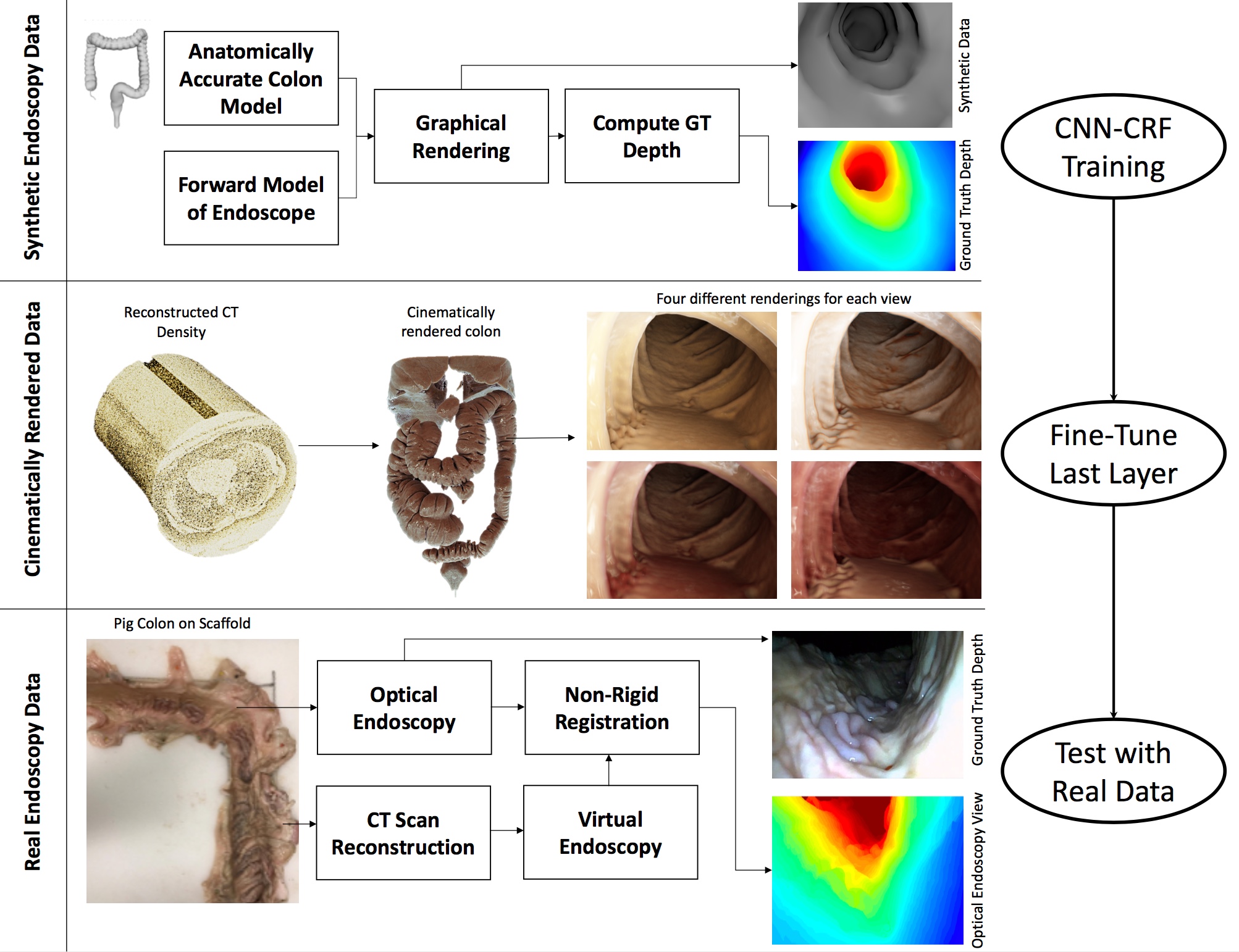}
\caption{Data generation process for (a) Synthetically generated data for training (b) Cinematically rendered data for fine-tuning and (c) Pig colon data for validation.}
\end{figure}

\subsection{Fine-Tuning Deep Networks}
\begin{figure}
\centering
\includegraphics[width=\textwidth]{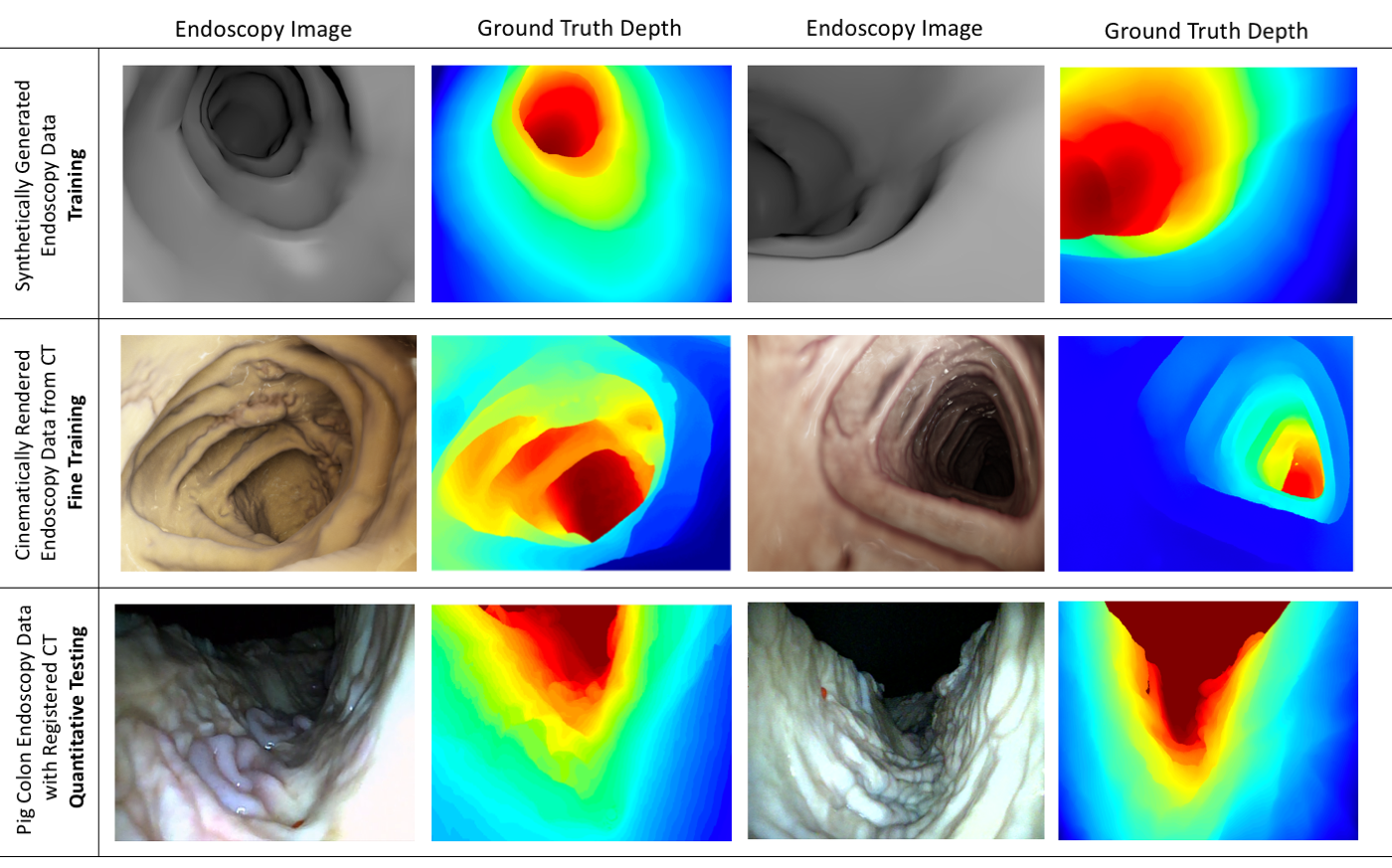}
\caption{Representative images of synthetically generated endoscopy data with ground truth depth for training (top), cinematically rendered CT data with ground truth depth for fine-tuning (middle) and real endoscopy data with ground truth depth from registered CT views for testing (bottom).}
\end{figure}

CNNs are trained by solving a typically non-convex error function using local search algorithms such as stochastic gradient descent and other optimizations. Beginning with randomly initialized weights, CNNs seek to minimize their empirical risk over the training dataset by iteratively updating the network parameters in the opposite direction of its error gradient, such that the network's performance converges towards a minimum on the loss surface \citep{bottou2010large}. With limited data, poor initialization, and a lack of regularization to control capacity, the network may fail to generalize, and convergence can become slow when traversing saddle points and also lead to sub-optimal local minima \citep{neyshabur2017exploring,choromanska2015loss,zhang2016understanding,dauphin2014identifying,samala2018evolutionary}. Initializing weights from a CNN trained for a similar task with a much larger dataset however, allows the network to converge much more easily to a good local minima and necessitates less labeled data \citep{glorot2010understanding}. { This process is called transfer learning, and is widely used in classification and segmentation tasks such as lesion detection in medical imaging, where there exists a paucity of annotated data \citep{penatti2015deep,azizpour2015generic,girshick2014rich,sonntag2017fine,zhen2017transfer,ravi2017transfer}. In practice, transfer learning involves transferring weights from an existing network trained on a much larger dataset. For networks trained on similar tasks and datasets, the new network would freeze the first few layers, and train the remaining layers at a low learning rate. This process is called fine-tuning. Intuitively, the first few layers of a CNN hold low-level features that are shared across all types of images, and the last layers hold high-level features that are learned for a specific application \citep{tajbakhsh2016convolutional,zhou2017fine,ravi2017transfer}.} In this specific context, we hypothesize that networks trained on synthetic medical data, that might not have previously adapted to real data, would generalize better if fine-tuned using cinematically rendered photorealistic data. We further hypothesize that such fine-tuned networks require less amount of training data, and would work well in low-resource settings such as endoscopy.


\subsection{Depth Estimation for Endoscopy}

For the purpose of validating our hypotheses we focus on the task of depth estimation from monocular endoscopy images. Monocular depth estimation from endoscopy is a challenging problem and has a variety of clinical applications including topographical reconstruction of the lumen, image-guided surgery, endoscopy quality metrics, and enhanced polyp detection, as polyps can lie on convex surfaces and can be occluded by folds in the gastrointestinal tract \citep{hazirbas2016fusenet,zhu2010polypCT,wang2015cadpolyp}. Depth estimation is especially challenging because the tissue being imaged is often deformable, and endoscopes have a single camera with close light sources and a wide field of view. Current approaches either have limited accuracy due to restrictive assumptions \citep{hong20143d} or require modifying endoscope hardware which has significant regulatory and engineering barriers \citep{parot2013photometric,durr2014pse}. Data-driven approaches for depth estimation in endoscopy are additionally complicated because of the lack of clinical images with available ground truth data, since it is difficult to include a depth sensor on an endoscope \citep{nadeem2016computer}. Moreover, networks trained on data from one patient fail to generalize to other patients since they start learning from patient-specific texture and color. Previous work has focused on generating synthetic data and adversarial domain adaptation to overcome these issues \citep{mahmood2017deep,mahmood2017unsupervised}. In this paper, we will focus on using synthetic endoscopy data with ground truth depth for training and fine-tuning using photorealistic cinematically rendered data.

\section{Methods}

\subsection{Endoscopy Depth Dataset Generation}

We generated three different datasets of endoscopy images with ground truth depth for three different purposes: (a) a large dataset of synthetic endoscopy images for training, (b) a small dataset of cinematically rendered images for fine-tuning, and (c) a small dataset of real endoscopy images of a porcine colon for validation.

\subsubsection{Synthetic Endoscopy Data for Training\\}

Though synthetic data has been extensively used to train deep CNN models for real-world images \citep{su2015render,gupta2016synthetic,varol2017learning,planche2017depthsynth}, this approach has been relatively limited for medical imaging.

Recent work in generating synthetic data for medical images have been applying GANs to retinal images and histopathology images \citep{costa2017end}. However, GAN-synthesized medical data does not cater for the cross-patient adaptability problem. In general, synthetic medical imaging data can be generated given an anatomically correct organ model and a forward model of an imaging device (Fig. 1-Top). Forward models for diagnostic imaging devices are more complicated than typical cameras and anatomic models of organs need to represent a high degree of variation. We developed a forward model of an endoscope with a wide-angle monocular camera and two to three light sources that exhibit realistic inverse square law of intensity fall-off. {We use a synthetically-generated and anatomically accurate colon model and image it using the virtual endoscope placed at a variety of angles and varying conditions to mimic the movement of an actual endoscope. We also generate pixel-wise ground truth depth for each rendered image. Using this model, we generated a dataset of 200,000 grayscale endoscopy images, each with a corresponding, error-free ground truth depth map (Fig. 1-Top, Fig. 2).}

\subsection{Cinematically Rendered Data for Fine-Tuning}

{While synthetic endoscopy data models the inverse of intensity fall off with depth, it conventionally only considers the surface of the rendered object\textemdash it does not simulate light scattering and extinction through turbid media. Moreover, conventional synthetic rendering does not simulate high frequency details in the colon such as texture and color. A model trained on data from one source is often incapable of performing well on a target domain due to distinctions in the distributions of these two domains. Models trained on synthetic data would have a domain bias towards synthetic images, and would not cover the distribution of testing cases found in real patient data. The Cinematic VRT technology developed at Siemens Healthcare provides a natural and photorealistic 3D representation of medical scans, such as Computed Tomography (CT) or Magnetic Resonance Images (MRI) \citep{comaniciu2016shaping,dappa2016cinematic}. The cinematic rendering process is computationally complex and depending on the image size it can take $5$ to $30$ seconds per image \citep{dappa2016cinematic}. Cinematic rendered data covers the testing use cases better than synthetic data. Our fine-tuning approach prevents the network from learning patient-specific features by assigning the same depth to four different cinematic renderings. By including renderings of the same colon image with different colors and textures in the training set, the network can learn more domain-invariant features, which would allow it to generalize well to other tissue models.}

The physical rendering algorithm, based on a Monte Carlo path-tracing technique closely simulates the complex interaction of light rays with tissues found in the scanned volume. Compared to traditional volume ray casting, where only light emission and absorption along a straight ray is considered, path tracing considers light paths with multiple random scattering events and light extinction.  Although this lighting model requires more computational power as hundreds of light paths must be calculated, it considerably  enhances  depth  and shape  perception.  By  putting  the  anatomical  structures  within  the medical  scans  in a  virtual  lighting  condition  that  mimics  the  physical  lighting  experienced  in  reality, soft shadows, ambient occlusions and volumetric scattering effects can be observed in the cinematic rendered images. Monte Carlo path tracing and interaction can be used to calculate the radiant flux, $L$ at a distance $x$ received from the direction $\omega$ along a ray using the following multidimensional rendering equation,

\begin{equation}\label{eq:Tk}
  \begin{aligned}
L(x,\omega)=\int_{0}^{D} e^{-\tau (x,x')} \sigma_s(x') \bigg[\int_{\Omega_{4\pi}}^{} p(\omega,\omega')L_i(x',\omega')d\omega'\bigg] d{x'},
  \end{aligned}
\end{equation}

\noindent{where, $\omega'$ represents all possible light directions and D represents the maximum distance. The optical properties of the tissue under consideration are defined by $p(\omega,\omega')$, which describes the fraction of light traveling along a direction $\omega'$ being scattered into direction $\omega. L_i(x',\omega')$} is the radiance arriving at distance $x'$ from direction $\omega'$. Surface interactions are modeled with a bidirectional reflectance distribution function (BRDF) and tissue scattering is modeled using a Henyey-Greenstein phase function \citep{toublanc1996henyey}. $\tau(x,x')=\int_{x}^{x'} \sigma_t(t) dt$ represents the optical depth and its corresponding excitation coefficient is represented by the sum of absorption and scattering coefficients, $\sigma_t=\sigma_s+\sigma_a$ \citep{comaniciu2016shaping}. Compared to conventional medical rendering, this technique considerably enhances depth and shape perception by putting the anatomical structures within the medical scans in a virtual lighting condition that mimics the physical lighting experienced in reality. Cinematic rendering has been used for a variety of medical imaging visualization tasks \citep{johnson2017mdct,rowe20183d,chu2018cinematic,rowe2018cinematic}.

Using this Cinematic VRT technology, colonic images were generated together with their corresponding depth maps, by saving the gradient and the position of the rays once their accumulated opacity had reached a given threshold (Fig. 1-Middle). Four different sets of rendering parameters were used to generate a diverse set of renderings for each scene. This was done to prevent the network from learning texture and color in the renderings (Fig. 1-Middle, Fig. 2). We used a total of 1200 rendered images for fine-tuning from 300 different scenes. {The CT colonoscopy data used was acquired from 9 patients from the NIH Cancer Imaging Archive (TCIA) \citep{johnson2008accuracy}.} Cinematic rendered data covers the testing use cases better than synthetic data, as it presents a possible solution to the domain adaptation problem with a more realistic forward model of the light-tissue interaction and by modeling both high frequency features and depth cues. Our fine-tuning approach prevents the network from learning patient-specific features by assigning the same depth to four different cinematic renderings. By including renderings of the same colon image with different colors and textures in the training set, the network can learn more domain-invariant features, which would allow it to generalize well to other tissue models. 

\subsection{Real Pig Colon Optical Endoscopy Data}

{To validate our approach, we tested the depth-estimation performance on a dataset of $1460$ real endoscopy images.} We created a dataset of ex-vivo pig colon optical endoscopy images with ground truth depth determined from CT. In particular, we fixed a porcine colon to a tubular scaffold and conducted optical endoscopy imaging using a Misumi Endoscope (MO-V5006L). Subsequently, we collected cone beam CT data from the same scaffold. A 3D model of this fixed colon was reconstructed using filtered-back projection with a Ram-Lak filter \citep{natterer1986mathematics}. The reconstructed density was then imaged using a virtual endoscope with same camera parameters as the optical Misumi endoscope. The resulting virtual endoscopy images were registered to optical endoscopy views using a one-plus-one evolutionary optimizer \citep{styner2000parametric,zitzler2004tutorial}. Once registered, the depth for each virtual endoscopy view was used as the depth for the corresponding optical endoscopy view (Fig.1-Bottom, Fig. 2).

\subsection{Monocular Endoscopy Depth Estimation using CNN-CRF Joint Training}

To train an endoscopy depth estimation network using synthetic data and fine-tuning using cinematically rendered data we used a joint CNN and Conditional Random Fields (CRF) network similar to the setup described in \citep{mahmood2017deep,liu_learning_2016}. Intuitively, a CNN-CRF setup is more context-aware than a simple CNN, as it takes into account the smooth transitions and abrupt changes that are characteristic of an endoscopy depth map. Assuming $\vec{g}\in \R^{n\times m}$ is an endoscopy image which has been divided into super-pixels, $p$, and $\vec{y}=[y_1,y_2,...,y_p] \in \R$ is the depth vector for each super-pixel. The conditional probability distribution of can be defined as,

\begin{equation}\label{eq:Tk}
  \begin{aligned}
 {Pr(\vec{y}|\vec{x})}=\frac{exp(E(\vec{y},\vec{x}))}{\int_{-\infty}^{\infty} exp(E(\vec{y},\vec{x})) d\vec{y}}.
  \end{aligned}
\end{equation}

 \noindent where, $E$ is the energy CRF function. In order to predict the depth of a new image we need to solve a maximum aposteriori (MAP) problem, $\widehat{\vec{y}}=\argmax_{y} {Pr(\vec{y}|\vec{x})}$. 

\noindent Let $\xi$ and $\eta$ be unary and pairwise potentials over nodes $\N$ and edges $\Sp$ of $\vec{x}$. Where, $\xi$ predicts the depth from a single superpixel and $\eta$ encourages smoothness between neighboring pairwise superpixels. The two potentials must be learned in a single unified framework. Based on \citep{liu_learning_2016,mahmood2017deep} the unary potential can be defined as, 

\begin{equation}\label{eq:Tk}
  \begin{aligned}
 \xi(y_{i},\vec{x};\vec\theta)= - (y_i-h_i(\vec{\theta}))^2
  \end{aligned}
\end{equation}

 \noindent where $h_i$ is the depth of a superpixel and $\theta$ represents CNN parameters. 
 \noindent The pairwise potential function is based on standard CRF vertex and edge feature functions studied extensively in \citep{qin_global_2009} and other works. Let $\vec{\beta}$ be the parameters of the network and $\vec{S}$ be the similarity index matrix where ${S}_{i,j}^k$ represents a metric between the $i^{th}$ and $j^{th}$ super-pixel. In this case, we used intensity and greyscale histogram as pairwise similarities which were expressed in $\ell_2$ form. The pairwise potential can then be simply written as,

\vspace{-1.0em}

\begin{equation}\label{eq:Tk}
  \begin{aligned}
 \eta(y_{i},y_{j};\vec\beta)= - \frac{1}{2}\sum_{k=1}^{K}\beta_{k}S_{i,j}^{k}(y_i-y_j)^2. 
  \end{aligned}
\end{equation}

\noindent Simplifying the energy function,

\vspace{-1.0em}

\begin{equation}\label{eq:Tk}
  \begin{aligned}
 E = - \sum_{i \in \N} (y_i-h_i(\vec{\theta}))^2  - \frac{1}{2}\sum_{(i,j) \in \Sp}\sum_{k=1}^{K}\beta_{k}S_{i,j}^{k}(y_i-y_j)^2.
  \end{aligned}
\end{equation}

\noindent During training, the negative log likelihood of the probability density function which can be derived from Eq. 1 is minimized with respect to the two learning parameters. Regularization terms $(\lambda_\theta,\lambda_\beta)$ are added to the objective function to suppress heavily weighted vectors. Let $N$ be the number of images in the training data then the objective function can be stated as,

\vspace{-1.0em}

\begin{equation}\label{eq:Tk}
  \begin{aligned}
  \min_{\theta,\beta \geq 0} {-\sum_{1}^{N}\log Pr(\vec{y}|\vec{x};\vec{\theta},\vec{\beta}})+ \frac{\lambda_\theta}{2} \norm{\theta}_2^2+ \frac{\lambda_\beta}{2} \norm{\beta}_2^2.
  \end{aligned}
\end{equation}

\noindent {This optimization problem is solved using stochastic gradient decent-based back propagation. Our network operates for the unary part on a $224\times224$ superpixel patch level and is composed of 5 convolutional and 4 fully connected layers (Fig. 3). The fully connected layers are fine-tuned using cinematically rendered data. The pairwise part operates on similarity $\ell_2$ similarity metrics between neighboring superpixels based on intensity and grayscale histogram followed by a fully connected layer. The network architecture is illustrated in Fig. 3. The network was initially trained on gray scale synthetic images with three channels, each channel assigned the same grayscale value. The network is then fine-tuned with color RGB cinematically rendered data. The network was trained using MatConvNet with Matlab 2017b. The momentum was set to 0.8 and the weight decay parameters were set to 0.0005. The network was trained for 200 epochs and the learning rate was set to 0.0002 and linearly decreased after the first 20 epochs. All parameters were tuned on synthetic data and cinematically rendered data, none of the real endoscopy test images were used for training or parameter tuning.}

\begin{figure}
\centering
\includegraphics[width=\textwidth]{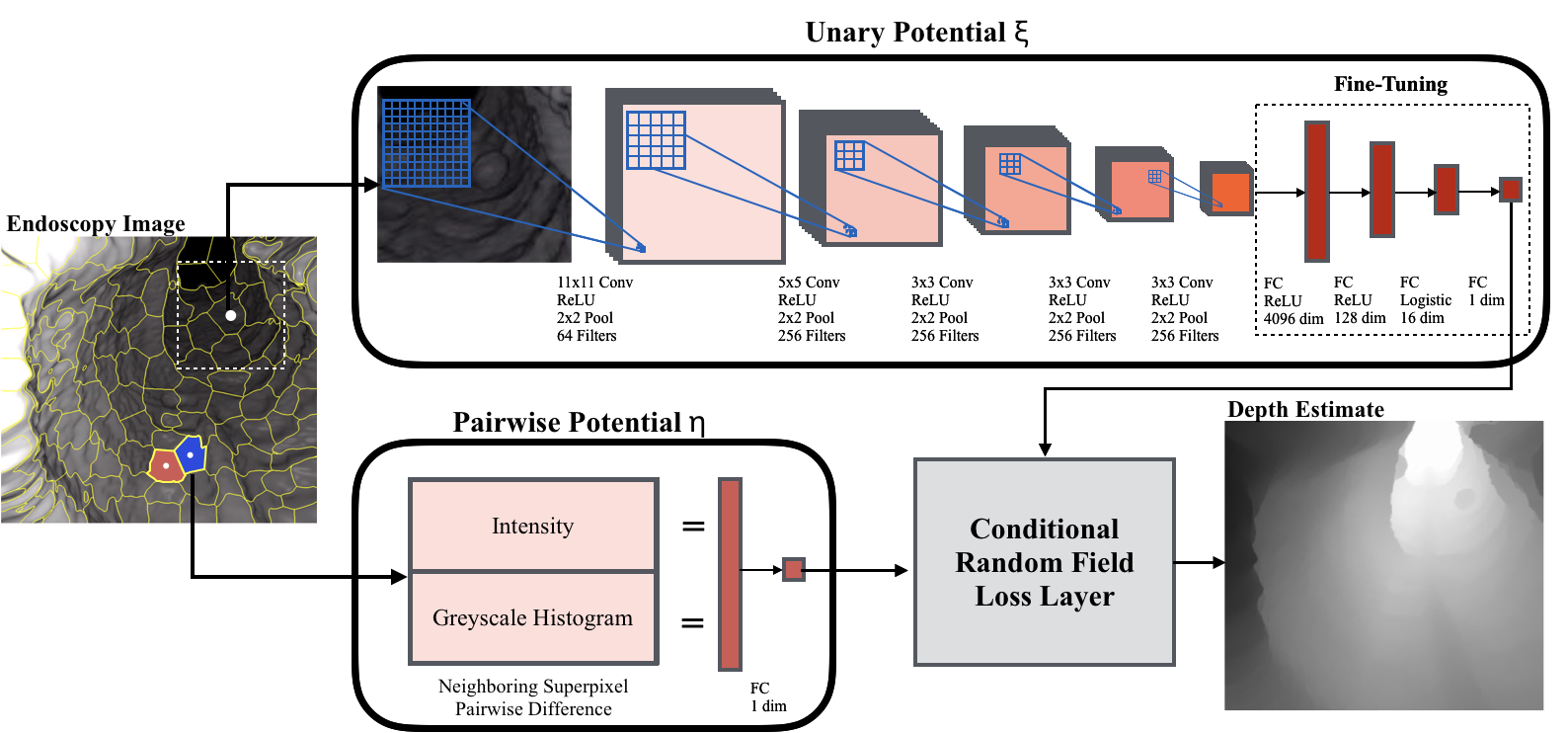}
\caption{Architecture of a CNN-CRF network with fine-tuning. The unary part is composed of 5 convolutional and 4 fully connected layers and the pairwise part is composed of a fully convolutional layer. Repeating units of this setup can be used for parallel processing.}
\end{figure}

\section{Results}

\subsection{Quantitative Evaluation}


We evaluated based on metrics our depth estimation paradigm and the capability of fine-tuning using the following metrics:

\begin{enumerate}
\item Relative Error (rel): $\frac{1}{N} \sum_{y} \frac{\lvert y_{gt}-y_{est} \rvert}{y_{gt}}$
\item Average $log_{10}$ Error ($log_{10}$): $\frac{1}{N} \sum_{y} \lvert \log_{10}y_{gt}-\log_{10}y_{est} \rvert$
\item Root Mean Square Error (rms):$\sqrt{\frac{1}{N} \sum_{y} (y_{gt}-y_{est})^2}$
\end{enumerate}

Where $y_{gt}$ is the ground truth depth $y_{est}$ is the estimated depth and $N$ is the total number of samples. Table 1 and 2 and Fig. 4 show results based on these metrics for cinematically data and porcine colon real endoscopy data. None of the test data was or images within the close proximity were used for training. Tables I and II validate our hypotheses that CNN-CRF fine-tuned networks (CNN-CRF-FT) works better than networks trained only on synthetic data and that a smaller amount of data is required for initial full training if the last layers are fine-tuned. {We also observed that fine-tuning with four renderings of a scene improved performance over fine-tuning with just one rendering (Table 2). This is because in supplying multiple renderings of the same scene with the same depth map, the CNN-CRF was able to better learn the context-aware features for depth estimation such as intensity differences between superpixels, and overcome noisy details such as texture or color.} As a result, the network was able to work well on real data such as the pig colon data used in this study. Table 3 shows that fine-tuning with just $300$ images from one kind of rendering gives a worse result compared to fine-tuning with $75$ images each from four different kinds of renderings. {Our experimentation demonstrated that training with only cinematically rendered data resulted in overfitting to high frequency features of the data. We found that training on synthetic data with lower frequency details and fine tuning on cinematically rendered data worked much better.}

\begin{figure}
\centering
\includegraphics[width=\textwidth]{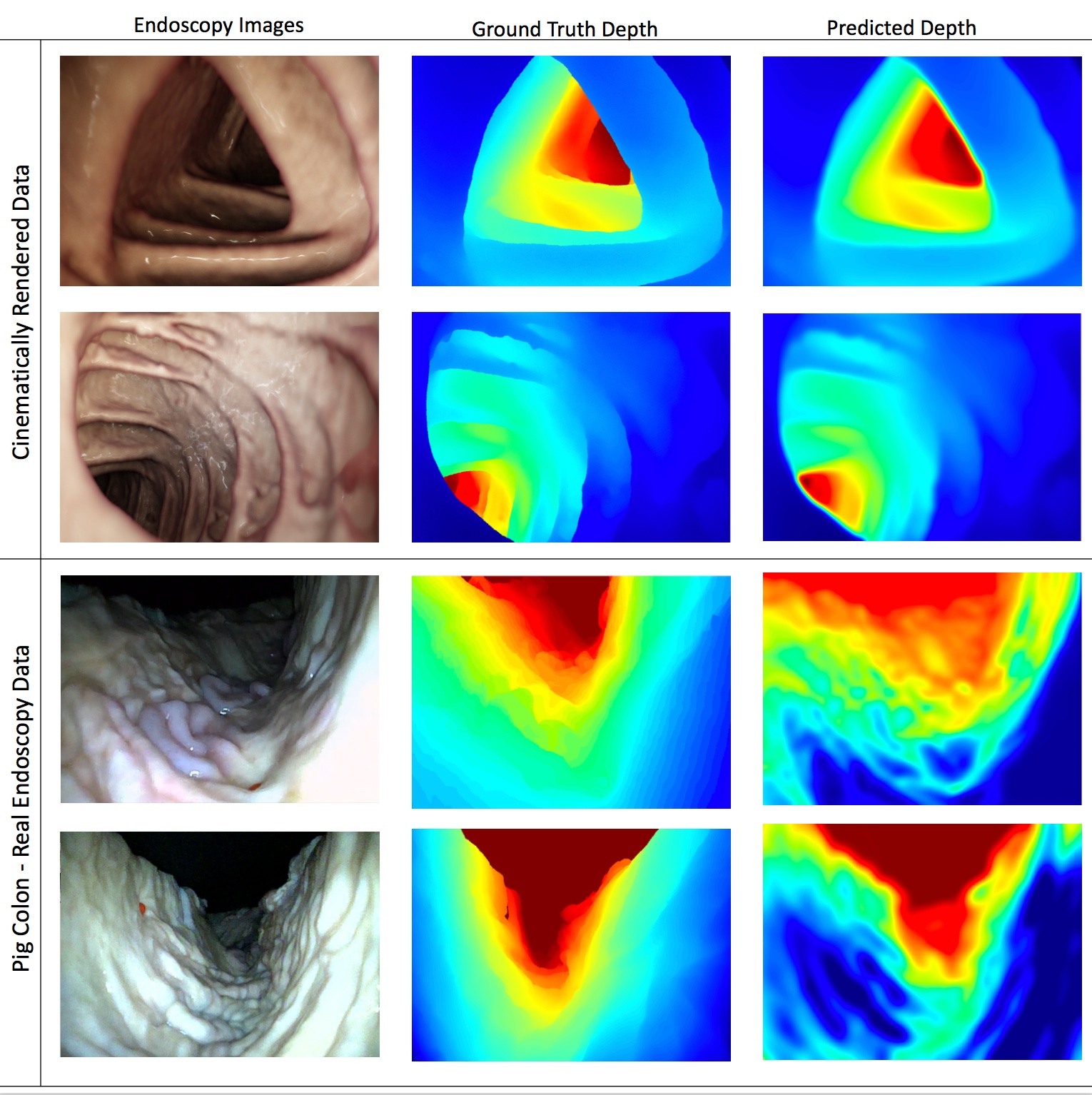}
\caption{Representative images of estimated depth and corresponding ground truth depth for cinematically rendered data and real pig colon endoscopy data. }
\end{figure}

\begin{table}[h]
\setlength{\tabcolsep}{7pt}

  \centering
  \begin{threeparttable}[b]
  \captionsetup{justification=centering,   textfont={sc}}
  \caption{Performance Evaluation for Cinematically Rendered Endoscopy Images}
  \label{tab:test2}
  \begin{tabular}{lllllll}
  \toprule
    \multicolumn{1}{c}{Method} & {Training} & {Fine-Tuning} &{rel $\downarrow$} & {$\log_{10}$ $\downarrow$} & {rms $\downarrow$}   \\
   \cmidrule(l){1-6}
   CNN-CRF		          & 200,000   &None        &0.394     &0.241    &1.933\\
   CNN-CRF-FT	          & 200,000   &1200        &\textbf{0.166}     &\textbf{0.078}    &\textbf{0.708}\\
   CNN-CRF      		  & 100,000   &None        &0.477     &0.319    &2.590\\
   CNN-CRF-FT			  & 100,000   &1200        &\textbf{0.189}     &\textbf{0.081}    &\textbf{0.756}\\

   \hline
  \end{tabular}
 \end{threeparttable}
\end{table}

\begin{table}[h]
\setlength{\tabcolsep}{7pt}

  \centering
  \begin{threeparttable}[b]
  \captionsetup{justification=centering,   textfont={sc}}
  \caption{Performance Evaluation for Real Pig Colon Endoscopy Images}
  \label{tab:test2}
  \begin{tabular}{lllllll}
  \toprule
    \multicolumn{1}{c}{Method} & {Training} & {Fine-Tuning} &{rel $\downarrow$} & {$\log_{10}$ $\downarrow$} & {rms $\downarrow$}   \\
   \cmidrule(l){1-6}
   CNN-CRF		          & 200,000   &None        &0.411     &0.258    &2.318\\
   CNN-CRF-FT	          & 200,000   &1200        &\textbf{0.298}     &\textbf{0.171}    &\textbf{1.387}\\
   CNN-CRF  			  & 100,000   &None        &0.495     &0.392    &2.897\\
   CNN-CRF-FT			  & 100,000   &1200        &\textbf{0.329}     &\textbf{0.196}    &\textbf{1.714}\\

   \hline
  \end{tabular}
 \end{threeparttable}
\end{table}

\begin{table}[h]
\setlength{\tabcolsep}{7pt}

  \centering
  \begin{threeparttable}[b]
  \captionsetup{justification=centering,   textfont={sc}}
  \caption{Performance Evaluation for Fine-Tuning with One vs Four Different Rendering (Pig Colon Data)}
  \label{tab:test2}
  \begin{tabular}{lllllll}
  \toprule
    \multicolumn{1}{c}{Method} & {Training} & {Fine-Tuning} &{rel $\downarrow$} & {$\log_{10}$ $\downarrow$} & {rms $\downarrow$}   \\
   \cmidrule(l){1-6}
   CNN-CRF-FT ($1\times300$ Rendering)	    & 200,000   &300     &0.471     &0.363    &2.614\\
   CNN-CRF-FT ($4\times75$ Renderings)      & 200,000   &300     &\textbf{0.364}     &\textbf{0.221}    &\textbf{2.153}\\

   \hline
  \end{tabular}
 \end{threeparttable}
\end{table}


\section{Conclusion}


{

Despite the recent advances in computer vision and deep learning algorithms, their applicability to medical images is often limited by the scarcity of annotated data. The problem is further complicated by the underrepresentation of rare conditions. For example, getting annotated data for polyp localization is difficult because in a 20 minute colonoscopy examination of the 1.5 meter colon, only a few 10 mm polyps may be present. Depending on the field-of-view of the camera, the polyps may also be occluded by folds in the gastrointestinal tract \citep{wang2015cadpolyp}. Using depth with RGB images has shown to improve localization in natural scenes by helping recover rich structural information with less annotated data and better cross-dataset adaptability \citep{hazirbas2016fusenet}. 

Within the constrained setting of endoscopy, estimating depth from monocular views is difficult because ground truth depth is hard to acquire. Problems where ground truth is difficult or impossible to acquire have been tackled for natural scenes by generating synthetic data. However, there are few examples of synthetic data driven medical imaging applications \citep{mahmood2017unsupervised,mahmood2017deep,nie2017medical}. This is because synthetic data-driven models often fail to generalize to the real datasets, and is often complimented by the cross patient network adaptability problem, where networks that work well on one patient do not generalize to other patients. 

{In this work, we demonstrate one of the first successful uses of cinematically rendered data for generalizing a network trained on synthetic data to real data. Additionally, our approach successfully addresses the issue of the domain adaptation or cross-patient network adaptability. We show that a synthetic data-driven CNN-CRF model can be successfully trained for accurate depth estimation on real tissue given no real optical endoscopy training data. Moreover, we prevent the network from learning from texture or color by providing the network with a variety of renderings assigned to the same depth value. We observed that depth estimation accuracy increases when trained with a variety of renderings for the same scene. We believe this improvement is due to the multiple rendering facilitating the networking learning to predict depth from cues that are invariant among different patients such as colon shape and light intensity fall-off with depth.}

{The shortcomings of this method include errors and artifacts in CT rendering. Unlike the synthetic data the ground truth for rendered CT data suffers from errors due to CT reconstruction artifacts, resolution limits, and imperfect registration. This error can be propagated in the learning process when fine-tuning with cinematically rendered data and when evaluating our model with real endoscopy data registered to CT.} 

Beyond accurate depth estimation, future work will investigate semantic segmentation in endoscopy by fusing depth as an additional input \citep{hazirbas2016fusenet}. Our future work will also focus on generalizing this concept to other medical imaging modalities.}

\section*{Acknowledgments}

The authors would like to thank Sermet Onel for his help with internal lighting and Kaloian Petkov for his help with multiple aspects of the cinematic renderer.

\section*{Disclaimer}

This feature is based on research, and is not commercially available. Due to regulatory reasons its future availability cannot be guaranteed.

\bibliographystyle{apalike}

\begin{thebibliography}{}

\bibitem[Azizpour et~al., 2015]{azizpour2015generic}
Azizpour, H., Razavian, A.~S., Sullivan, J., Maki, A., and Carlsson, S. (2015).
\newblock From generic to specific deep representations for visual recognition.
\newblock In {\em CVPRW DeepVision Workshop, June 11, 2015, Boston, MA, USA}.
  IEEE conference proceedings.

\bibitem[Bottou, 2010]{bottou2010large}
Bottou, L. (2010).
\newblock Large-scale machine learning with stochastic gradient descent.
\newblock In {\em Proceedings of COMPSTAT'2010}, pages 177--186. Springer.

\bibitem[Choromanska et~al., 2015]{choromanska2015loss}
Choromanska, A., Henaff, M., Mathieu, M., Arous, G.~B., and LeCun, Y. (2015).
\newblock The loss surfaces of multilayer networks.
\newblock In {\em Artificial Intelligence and Statistics}, pages 192--204.

\bibitem[Chu et~al., 2018]{chu2018cinematic}
Chu, L.~C., Johnson, P.~T., and Fishman, E.~K. (2018).
\newblock Cinematic rendering of pancreatic neoplasms: preliminary observations
  and opportunities.
\newblock {\em Abdominal Radiology}, pages 1--7.

\bibitem[Comaniciu et~al., 2016]{comaniciu2016shaping}
Comaniciu, D., Engel, K., Georgescu, B., and Mansi, T. (2016).
\newblock Shaping the future through innovations: From medical imaging to
  precision medicine.
\newblock {\em Medical image analysis}, 33:19--26.

\bibitem[Costa et~al., 2017]{costa2017end}
Costa, P., Galdran, A., Meyer, M.~I., Niemeijer, M., Abr{\`a}moff, M.,
  Mendon{\c{c}}a, A.~M., and Campilho, A. (2017).
\newblock End-to-end adversarial retinal image synthesis.
\newblock {\em IEEE Transactions on Medical Imaging}.

\bibitem[Creswell et~al., 2018]{creswell2017gan}
Creswell, A., White, T., Dumoulin, V., Arulkumaran, K., Sengupta, B., and
  Bharath, A.~A. (2018).
\newblock Generative adversarial networks: An overview.
\newblock In {\em IEEE Signal Processing Magazine}, volume~35, pages 53--65.

\bibitem[Dappa et~al., 2016]{dappa2016cinematic}
Dappa, E., Higashigaito, K., Fornaro, J., Leschka, S., Wildermuth, S., and
  Alkadhi, H. (2016).
\newblock Cinematic rendering--an alternative to volume rendering for 3d
  computed tomography imaging.
\newblock {\em Insights into imaging}, 7(6):849--856.

\bibitem[Dauphin et~al., 2014]{dauphin2014identifying}
Dauphin, Y.~N., Pascanu, R., Gulcehre, C., Cho, K., Ganguli, S., and Bengio, Y.
  (2014).
\newblock Identifying and attacking the saddle point problem in
  high-dimensional non-convex optimization.
\newblock In {\em Advances in neural information processing systems}, pages
  2933--2941.

\bibitem[Durr et~al., 2014a]{durr2014pse}
Durr, N.~J., González, G., Lim, D., and Traverso, G. (2014a).
\newblock Endoscopic-ct: learning-based photometric reconstruction for
  endoscopic sinus surgery.
\newblock In {\em Advanced Biomedical and Clinical Diagnostic Systems}, volume
  8935. International Society for Optics and Photonics.

\bibitem[Durr et~al., 2014b]{durr2014colon}
Durr, N.~J., González, G., and Parot, V. (2014b).
\newblock 3d imaging techniques for improved colonoscopy.
\newblock {\em Expert Review of Medical Devices}, 11(2):105--107.

\bibitem[Eid et~al., 2017]{eid2017cr}
Eid, M., Cecco, C. N.~D., Nance, J.~W., Jr., Caruso, D., Albrecht, M.~H.,
  Spandorfer, A.~J., Santis, D.~D., Varga-Szemes, A., and Schoepf, U.~J.
  (2017).
\newblock Cinematic rendering in ct: A novel, lifelike 3d visualization
  technique.
\newblock {\em American Journal of Roentgenology}, 209(2).

\bibitem[Girshick et~al., 2014]{girshick2014rich}
Girshick, R., Donahue, J., Darrell, T., and Malik, J. (2014).
\newblock Rich feature hierarchies for accurate object detection and semantic
  segmentation.
\newblock In {\em Proceedings of the IEEE conference on computer vision and
  pattern recognition}, pages 580--587.

\bibitem[Glorot and Bengio, 2010]{glorot2010understanding}
Glorot, X. and Bengio, Y. (2010).
\newblock Understanding the difficulty of training deep feedforward neural
  networks.
\newblock In {\em Proceedings of the thirteenth international conference on
  artificial intelligence and statistics}, pages 249--256.

\bibitem[Goodfellow et~al., 2016]{goodfellow2016deep}
Goodfellow, I., Bengio, Y., Courville, A., and Bengio, Y. (2016).
\newblock {\em Deep learning}, volume~1.
\newblock MIT press Cambridge.

\bibitem[Goodfellow et~al., 2014]{goodfellow2014generative}
Goodfellow, I., Pouget-Abadie, J., Mirza, M., Xu, B., Warde-Farley, D., Ozair,
  S., Courville, A., and Bengio, Y. (2014).
\newblock Generative adversarial nets.
\newblock In {\em Advances in neural information processing systems}, pages
  2672--2680.

\bibitem[Greenspan et~al., 2016]{greenspan2016guest}
Greenspan, H., van Ginneken, B., and Summers, R.~M. (2016).
\newblock Guest editorial deep learning in medical imaging: Overview and future
  promise of an exciting new technique.
\newblock {\em IEEE Transactions on Medical Imaging}, 35(5):1153--1159.

\bibitem[Gupta et~al., 2016]{gupta2016synthetic}
Gupta, A., Vedaldi, A., and Zisserman, A. (2016).
\newblock Synthetic data for text localisation in natural images.
\newblock In {\em Proceedings of the IEEE Conference on Computer Vision and
  Pattern Recognition}, pages 2315--2324.

\bibitem[Gur et~al., 2017]{gur2017towards}
Gur, Y., Moradi, M., Bulu, H., Guo, Y., Compas, C., and Syeda-Mahmood, T.
  (2017).
\newblock Towards an efficient way of building annotated medical image
  collections for big data studies.
\newblock In {\em Intravascular Imaging and Computer Assisted Stenting, and
  Large-Scale Annotation of Biomedical Data and Expert Label Synthesis}, pages
  87--95. Springer.

\bibitem[Hazirbas et~al., 2016]{hazirbas2016fusenet}
Hazirbas, C., Ma, L., Domokos, C., and Cremers, D. (2016).
\newblock Fusenet: Incorporating depth into semantic segmentation via
  fusion-based cnn architecture.
\newblock In {\em Asian Conference on Computer Vision}, pages 213--228.
  Springer.

\bibitem[Hong et~al., 2014]{hong20143d}
Hong, D., Tavanapong, W., Wong, J., Oh, J., and De~Groen, P.~C. (2014).
\newblock 3d reconstruction of virtual colon structures from colonoscopy
  images.
\newblock {\em Computerized Medical Imaging and Graphics}, 38(1):22--33.

\bibitem[Johnson et~al., 2008]{johnson2008accuracy}
Johnson, C.~D., Chen, M.-H., Toledano, A.~Y., Heiken, J.~P., Dachman, A., Kuo,
  M.~D., Menias, C.~O., Siewert, B., Cheema, J.~I., Obregon, R.~G., et~al.
  (2008).
\newblock Accuracy of ct colonography for detection of large adenomas and
  cancers.
\newblock {\em New England Journal of Medicine}, 359(12):1207--1217.

\bibitem[Johnson et~al., 2017]{johnson2017mdct}
Johnson, P.~T., Schneider, R., Lugo-Fagundo, C., Johnson, M.~B., and Fishman,
  E.~K. (2017).
\newblock Mdct angiography with 3d rendering: a novel cinematic rendering
  algorithm for enhanced anatomic detail.
\newblock {\em American Journal of Roentgenology}, 209(2):309--312.

\bibitem[Kerkhof et~al., 2007]{kerkhof2007dysplasia}
Kerkhof, M., Van~Dekken, H., Steyerberg, E., Meijer, G., Mulder, A.,
  De~Bruïne, A., Driessen, A., Ten~Kate, F., Kusters, J., Kuipers, E., and
  Siersema, P. (2007).
\newblock Grading of dysplasia in barrett's oesophagus: substantial
  interobserver variation between general and gastrointestinal pathologists.
\newblock {\em Histopathology}, 50:920--927.

\bibitem[LeCun et~al., 2015]{lecun2015deep}
LeCun, Y., Bengio, Y., and Hinton, G. (2015).
\newblock Deep learning.
\newblock {\em nature}, 521(7553):436.

\bibitem[Liu et~al., 2016]{liu_learning_2016}
Liu, F., Shen, C., Lin, G., and Reid, I. (2016).
\newblock Learning depth from single monocular images using deep convolutional
  neural fields.
\newblock {\em IEEE transactions on pattern analysis and machine intelligence},
  38(10):2024--2039.

\bibitem[Mahmood et~al., 2018]{mahmood2017unsupervised}
Mahmood, F., Chen, R., and Durr, N.~J. (2018).
\newblock Unsupervised reverse domain adaption for synthetic medical images via
  adversarial training.
\newblock {\em IEEE Transactions on Medical Imaging}.

\bibitem[Mahmood and Durr, 2018a]{mahmood2017deep}
Mahmood, F. and Durr, N.~J. (2018a).
\newblock Deep learning and conditional random fields-based depth estimation
  and topographical reconstruction from conventional endoscopy.
\newblock {\em Medical Image Analysis}.

\bibitem[Mahmood and Durr, 2018b]{mahmood2018deep}
Mahmood, F. and Durr, N.~J. (2018b).
\newblock Deep learning-based depth estimation from a synthetic endoscopy image
  training set.
\newblock In {\em Medical Imaging 2018: Image Processing}, volume 10574, page
  1057421. International Society for Optics and Photonics.

\bibitem[Moradi et~al., 2016]{moradi2016cross}
Moradi, M., Guo, Y., Gur, Y., Negahdar, M., and Syeda-Mahmood, T. (2016).
\newblock A cross-modality neural network transform for semi-automatic medical
  image annotation.
\newblock In {\em International Conference on Medical Image Computing and
  Computer-Assisted Intervention}, pages 300--307. Springer.

\bibitem[Nadeem and Kaufman, 2016]{nadeem2016computer}
Nadeem, S. and Kaufman, A. (2016).
\newblock Computer-aided detection of polyps in optical colonoscopy images.
\newblock In {\em SPIE Medical Imaging}, pages 978525--978525. International
  Society for Optics and Photonics.

\bibitem[Natterer, 1986]{natterer1986mathematics}
Natterer, F. (1986).
\newblock {\em The mathematics of computerized tomography}, volume~32.
\newblock Siam.

\bibitem[Neyshabur et~al., 2017]{neyshabur2017exploring}
Neyshabur, B., Bhojanapalli, S., McAllester, D., and Srebro, N. (2017).
\newblock Exploring generalization in deep learning.
\newblock In {\em Advances in Neural Information Processing Systems}, pages
  5949--5958.

\bibitem[Nie et~al., 2017]{nie2017medical}
Nie, D., Trullo, R., Lian, J., Petitjean, C., Ruan, S., Wang, Q., and Shen, D.
  (2017).
\newblock Medical image synthesis with context-aware generative adversarial
  networks.
\newblock In {\em International Conference on Medical Image Computing and
  Computer-Assisted Intervention}, pages 417--425. Springer.

\bibitem[Parot et~al., 2013]{parot2013photometric}
Parot, V., Lim, D., Gonz{\'a}lez, G., Traverso, G., Nishioka, N.~S., Vakoc,
  B.~J., and Durr, N.~J. (2013).
\newblock Photometric stereo endoscopy.
\newblock {\em Journal of biomedical optics}, 18(7):076017.

\bibitem[Penatti et~al., 2015]{penatti2015deep}
Penatti, O.~A., Nogueira, K., and dos Santos, J.~A. (2015).
\newblock Do deep features generalize from everyday objects to remote sensing
  and aerial scenes domains?
\newblock In {\em Computer Vision and Pattern Recognition Workshops (CVPRW),
  2015 IEEE Conference on}, pages 44--51. IEEE.

\bibitem[Planche et~al., 2017]{planche2017depthsynth}
Planche, B., Wu, Z., Ma, K., Sun, S., Kluckner, S., Chen, T., Hutter, A.,
  Zakharov, S., Kosch, H., and Ernst, J. (2017).
\newblock Depthsynth: Real-time realistic synthetic data generation from cad
  models for 2.5 d recognition.
\newblock {\em arXiv preprint arXiv:1702.08558}.

\bibitem[Qin et~al., 2009]{qin_global_2009}
Qin, T., Liu, T.-Y., Zhang, X.-D., Wang, D.-S., and Li, H. (2009).
\newblock Global ranking using continuous conditional random fields.
\newblock In {\em Advances in neural information processing systems}, pages
  1281--1288.

\bibitem[Reiter et~al., 2016]{reiter2016endoscopic}
Reiter, A., L{\'e}onard, S., Sinha, A., Ishii, M., Taylor, R.~H., and Hager,
  G.~D. (2016).
\newblock Endoscopic-ct: learning-based photometric reconstruction for
  endoscopic sinus surgery.
\newblock In {\em Medical Imaging 2016: Image Processing}, volume 9784, page
  978418. International Society for Optics and Photonics.

\bibitem[Rowe et~al., 2018a]{rowe2018cinematic}
Rowe, S.~P., Chu, L.~C., and Fishman, E.~K. (2018a).
\newblock Cinematic rendering of small bowel pathology: preliminary
  observations from this novel 3d ct visualization method.
\newblock {\em Abdominal Radiology}, pages 1--10.

\bibitem[Rowe et~al., 2018b]{rowe20183d}
Rowe, S.~P., Zinreich, S.~J., and Fishman, E.~K. (2018b).
\newblock 3d cinematic rendering of the calvarium, maxillofacial structures,
  and skull base: preliminary observations.
\newblock {\em The British journal of radiology}, 91(xxxx):20170826.

\bibitem[Samala et~al., 2017]{ravi2017transfer}
Samala, R.~K., Chan, H.-P., Hadjiiski, L.~M., Helvie, M.~A., Cha, K.~H., and
  Richter, C.~D. (2017).
\newblock Multi-task transfer learning deep convolutional neural network:
  application to computer-aided diagnosis of breast cancer on mammograms.
\newblock {\em Physics in Medicine \& Biology}, 62(23):8894.

\bibitem[Samala et~al., 2018]{samala2018evolutionary}
Samala, R.~K., Chan, H.-P., Hadjiiski, L.~M., Helvie, M.~A., Richter, C., and
  Cha, K. (2018).
\newblock Evolutionary pruning of transfer learned deep convolutional neural
  network for breast cancer diagnosis in digital breast tomosynthesis.
\newblock {\em Physics in medicine and biology}.

\bibitem[Schlegl et~al., 2017]{schlegl2017unsupervised}
Schlegl, T., Seeb{\"o}ck, P., Waldstein, S.~M., Schmidt-Erfurth, U., and Langs,
  G. (2017).
\newblock Unsupervised anomaly detection with generative adversarial networks
  to guide marker discovery.
\newblock In {\em International Conference on Information Processing in Medical
  Imaging}, pages 146--157. Springer.

\bibitem[Shen et~al., 2017]{shen2017deep}
Shen, D., Wu, G., and Suk, H.-I. (2017).
\newblock Deep learning in medical image analysis.
\newblock {\em Annual Review of Biomedical Engineering}, (0).

\bibitem[Shin et~al., 2016]{shin2016deep}
Shin, H.-C., Roth, H.~R., Gao, M., Lu, L., Xu, Z., Nogues, I., Yao, J.,
  Mollura, D., and Summers, R.~M. (2016).
\newblock Deep convolutional neural networks for computer-aided detection: Cnn
  architectures, dataset characteristics and transfer learning.
\newblock {\em IEEE Transactions on Medical Imaging}, 35(5):1285--1298.

\bibitem[Sonntag et~al., 2017]{sonntag2017fine}
Sonntag, D., Barz, M., Zacharias, J., Stauden, S., Rahmani, V., F{\'o}thi,
  {\'A}., and L{\H{o}}rincz, A. (2017).
\newblock Fine-tuning deep cnn models on specific ms coco categories.
\newblock {\em arXiv preprint arXiv:1709.01476}.

\bibitem[Styner et~al., 2000]{styner2000parametric}
Styner, M., Brechbuhler, C., Szckely, G., and Gerig, G. (2000).
\newblock Parametric estimate of intensity inhomogeneities applied to mri.
\newblock {\em IEEE transactions on medical imaging}, 19(3):153--165.

\bibitem[Su et~al., 2015]{su2015render}
Su, H., Qi, C.~R., Li, Y., and Guibas, L. (2015).
\newblock Render for cnn: Viewpoint estimation in images using cnns trained
  with rendered 3d model views.
\newblock In {\em Proceedings of the IEEE International Conference on Computer
  Vision}, pages 2686--2694.

\bibitem[Tajbakhsh et~al., 2016]{tajbakhsh2016convolutional}
Tajbakhsh, N., Shin, J.~Y., Gurudu, S.~R., Hurst, R.~T., Kendall, C.~B.,
  Gotway, M.~B., and Liang, J. (2016).
\newblock Convolutional neural networks for medical image analysis: Full
  training or fine tuning?
\newblock {\em IEEE transactions on medical imaging}, 35(5):1299--1312.

\bibitem[Toublanc, 1996]{toublanc1996henyey}
Toublanc, D. (1996).
\newblock Henyey--greenstein and mie phase functions in monte carlo radiative
  transfer computations.
\newblock {\em Applied optics}, 35(18):3270--3274.

\bibitem[Varol et~al., 2017]{varol2017learning}
Varol, G., Romero, J., Martin, X., Mahmood, N., Black, M., Laptev, I., and
  Schmid, C. (2017).
\newblock Learning from synthetic humans.
\newblock {\em arXiv preprint arXiv:1701.01370}.

\bibitem[Wang et~al., 2015]{wang2015cadpolyp}
Wang, H., Liang, Z., Li, L.~C., Han, H., Song, B., Pickhardt, P.~J., Barish,
  M.~A., and Lascarides, C.~E. (2015).
\newblock An adaptive paradigm for computer-aided detection of colonic polyps.
\newblock {\em Physics in Medicine \& Biology}, 60(18):7207.

\bibitem[Wong et~al., 2017]{wong2017building}
Wong, K.~C., Karargyris, A., Syeda-Mahmood, T., and Moradi, M. (2017).
\newblock Building disease detection algorithms with very small numbers of
  positive samples.
\newblock In {\em International Conference on Medical Image Computing and
  Computer-Assisted Intervention}, pages 471--479. Springer.

\bibitem[Zhang et~al., 2016]{zhang2016understanding}
Zhang, C., Bengio, S., Hardt, M., Recht, B., and Vinyals, O. (2016).
\newblock Understanding deep learning requires rethinking generalization.
\newblock {\em arXiv preprint arXiv:1611.03530}.

\bibitem[Zhang et~al., 2017]{zhang2017deep}
Zhang, Y., Yang, L., Chen, J., Fredericksen, M., Hughes, D.~P., and Chen, D.~Z.
  (2017).
\newblock Deep adversarial networks for biomedical image segmentation utilizing
  unannotated images.
\newblock In {\em International Conference on Medical Image Computing and
  Computer-Assisted Intervention}, pages 408--416. Springer.

\bibitem[Zhang and Yuille, 2016]{zhang2016unreal}
Zhang, Y. and Yuille, A.~L. (2016).
\newblock Unrealstereo: Synthetic dataset for analyzing stereo vision.
\newblock {\em arXiv preprint arXiv:1612.04647}.

\bibitem[Zhen et~al., 2017]{zhen2017transfer}
Zhen, X., Chen, J., Zhong, Z., Hrycushko, B., Zhou, L., Jiang, S., Albuquerque,
  K., and Gu, X. (2017).
\newblock Deep convolutional neural network with transfer learning for rectum
  toxicity prediction in cervical cancer radiotherapy: a feasibility study.
\newblock {\em Physics in Medicine \& Biology}, 62(21):8246.

\bibitem[Zhou et~al., 2017]{zhou2017fine}
Zhou, Z., Shin, J., Zhang, L., Gurudu, S., Gotway, M., and Liang, J. (2017).
\newblock Fine-tuning convolutional neural networks for biomedical image
  analysis: actively and incrementally.
\newblock In {\em IEEE conference on computer vision and pattern recognition,
  Hawaii}, pages 7340--7349.

\bibitem[Zhu et~al., 2010]{zhu2010polypCT}
Zhu, H., Fan, Y., Lu, H., and Liang, Z. (2010).
\newblock Improving initial polyp candidate extraction for ct colonography.
\newblock {\em Physics in Medicine \& Biology}, 55(7):2087.

\bibitem[Zitzler et~al., 2004]{zitzler2004tutorial}
Zitzler, E., Laumanns, M., and Bleuler, S. (2004).
\newblock A tutorial on evolutionary multiobjective optimization.
\newblock In {\em Metaheuristics for multiobjective optimisation}, pages 3--37.
  Springer.

\end{thebibliography}

\end{document}